\documentclass[10pt,twocolumn,letterpaper]{article}

\usepackage{cvpr}              
\usepackage{url}
\usepackage{multirow}
\usepackage[ruled,vlined]{algorithm2e}
\usepackage{amsmath,bm}
\usepackage{amsfonts}
\usepackage{amssymb}
\usepackage{amsthm}

\usepackage{wrapfig,lipsum,booktabs}

\renewcommand\footnotemark{}

\usepackage[dvipsnames]{xcolor}

\usepackage{graphicx}
\usepackage{amsmath}
\usepackage{amssymb}
\usepackage{multirow}
\usepackage{booktabs}

\usepackage{wrapfig}

\definecolor{cvprblue}{rgb}{0.21,0.49,0.74}
\usepackage[pagebackref,breaklinks,colorlinks,citecolor=cvprblue]{hyperref}

\def\paperID{2935} 
\def\confName{CVPR}
\def\confName{CVPR}
\def\confYear{2024}

\title{LLMs are Good Sign Language Translators}

\author{Jia Gong\textsuperscript{1\dag}\thanks{ \dag~Equal contribution; 
 \ddag~Corresponding author}
~~~ Lin Geng Foo\textsuperscript{1\dag}
~~~ Yixuan He\textsuperscript{1\dag}
~~~ Hossein Rahmani\textsuperscript{2} ~~~ Jun Liu\textsuperscript{1\ddag} \\
\textsuperscript{1}Singapore University of Technology and Design ~~ 
\textsuperscript{2}Lancaster University\\
{\tt\small \{jia\_gong,lingeng\_foo,yixuan\_he\}@mymail.sutd.edu.sg,}\\ 
{\tt\small h.rahmani@lancaster.ac.uk, jun\_liu@sutd.edu.sg } \\
}

\begin{document}
\maketitle
\begin{abstract}
Sign Language Translation (SLT) is a challenging task that aims to translate sign videos into spoken language. Inspired by the strong translation capabilities of large language models (LLMs) that are trained on extensive multilingual text corpora, we aim to harness off-the-shelf LLMs to handle SLT. In this paper, we regularize the sign videos to embody linguistic characteristics of spoken language, and propose a novel SignLLM framework to transform sign videos into a language-like representation for improved readability by off-the-shelf LLMs. SignLLM comprises two key modules: (1) The Vector-Quantized Visual Sign module converts sign videos into a sequence of discrete character-level sign tokens, and (2) the Codebook Reconstruction and Alignment module converts these character-level tokens into word-level sign representations using an optimal transport formulation. A sign-text alignment loss further bridges the gap between sign and text tokens, enhancing semantic compatibility. We achieve state-of-the-art gloss-free results on two widely-used SLT benchmarks.
\end{abstract}

\section{Introduction}

Sign languages, which are visual signals expressed through hand, body, and facial movements, serve as the primary means of communication within the hearing-impaired community.
In an effort to facilitate effective communication with this community, much attention has been directed towards developing techniques to tackle the Sign Language Translation (SLT) task \cite{camgoz2018neural,camgoz2020sign,SLT:glossfree,SLT:SignBT,SLT:SLTUNET}, where the goal is to translate sign videos into spoken language.
SLT is a challenging task that requires cross-modality understanding of visual and linguistic cues \cite{SLT:glossfree,SLT:SignBT,SLT:SLTUNET}, and the challenge is exacerbated by the limited availability of paired sign-text data \cite{camgoz2020sign,SLT:SLTUNET,yin2020better,SLT:glossfree,SLT:MMTLB}.
Despite the notable advancements in terms of network architectures \cite{camgoz2018neural,camgoz2020sign,li2020tspnet}, visual sign representations (e.g., with keypoint estimators \cite{SLT:TwoStream,SLT:STMC-T}), and training methods \cite{SLT:glossfree,fu2023token,SLT:SLTUNET}, how to effectively tackle the challenging cross-modal SLT task with limited paired sign-text data largely remains an open question.

On the other hand, large language models (LLMs) -- referring to language models that have been trained on a large web-scale text corpus -- have recently received a lot of attention.
Since LLMs are trained over a very large corpus across multiple languages with distinct syntax and lexicon, they possess rich semantic understanding and powerful linguistic abilities
\cite{chowdhery2022palm,touvron2023llama,brown2020language}.
At the same time, LLMs have also demonstrated an impressive capability to translate across multiple languages \cite{chowdhery2022palm,brown2020language}, even showing a strong potential for translating languages with limited data \cite{yang2023bigtrans,zhu2023multilingual}. 
The foundation for this translation proficiency lies in the shared linguistic properties of syntax, lexicon, and morphology that many languages hold, which is particularly evident within language families  \cite{ammon2010world}.
Therefore, when faced with a new language with limited data, LLMs can draw upon the wealth of knowledge acquired from previously learned languages, leveraging any shared properties of syntax, lexicon and morphology with previous languages to effectively generate translations for new languages with remarkable accuracy and fluency \cite{yang2023bigtrans, zhu2023multilingual}.

Inspired by the impressive translation capabilities of LLMs, we aim to harness off-the-shelf LLMs to handle the challenging SLT task.
However, training LLMs directly on the relatively small SLT dataset can potentially lead to forgetting of their rich knowledge \cite{chen2020recall,he2021effectiveness} and a decline in performance, thus we follow previous LLM-based works \cite{melnyk2023reprog,gupta2023visual,song2023llm} to keep the off-the-shelf LLM frozen, which preserves the rich knowledge acquired during its pre-training on a vast multilingual corpus.
Consequently, our focus shifts towards making the sign videos compatible and readable for the off-the-shelf and frozen LLM to perform SLT. 
Specifically, in this paper, we explore the following question: \textit{Can we treat the sign video as a form of language, and leverage an off-the-shelf and frozen LLM to translate them?}
Notably, this is not a straightforward task because directly encoding features from sign videos with a pre-trained feature extractor \cite{he2016deep,wang2020deep} will result in a large gap between the sign video features and text tokens, making it difficult for off-the-shelf LLMs to understand them.

Based on the observation that LLMs can effectively handle new languages by leveraging shared commonalities with previously learned languages, we aim to introduce designs that transform our sign videos into a language-like format which are readable and friendly to LLMs.
Specifically, we hypothesize that providing language-like representations of sign videos to the LLM improves the LLM's understanding of the sign videos and facilitates greater exploitation of shared properties with previously learned languages, thus resulting in better SLT performance by the LLM.
To obtain language-like sign video representations, we draw inspiration from linguistic studies and analyses on LLMs \cite{vulic2020probing, hofmann2020dagobert} and regularize the sign video to embody two fundamental language-like characteristics:
\textbf{Discrete Characteristics:} Spoken languages are inherently discrete, since each language contains a finite set of words (and subwords) that convey distinct concepts, allowing them to be naturally represented through a discrete vocabulary with distinct tokens \cite{van2017neural,bostrom2020byte}. 
\textbf{Hierarchical Structure}: Most spoken languages exhibit three hierarchical semantic levels -- the sentence, word, and character levels \cite{levelt1999producing,seneff1998use}. 
This hierarchical structure enables languages to express a wide range of words with a limited set of characters, and convey diverse sentences with a limited number of words.

In this paper, we present SignLLM, a novel framework designed to regularize input sign videos to produce sign token representations with language-like characteristics that are compatible and friendly to LLMs. 
Our proposed SignLLM includes two key designs to impart discrete characteristics and a hierarchical structure to the produced sign tokens. 
Firstly, we introduce the Vector-Quantized Visual Sign (\textbf{VQ-Sign}) module that facilitates the conversion of sign videos into a \textit{sequence of discrete character-level sign tokens}.
To achieve this, the VQ-Sign module consists of a discrete character-level sign codebook which is optimized through a self-supervised context prediction task.
{Next, we introduce the Codebook Reconstruction and Alignment (\textbf{CRA}) module that converts the character-level sign tokens into \textit{word-level sign tokens}, facilitated by an optimal transport formulation.}
Moreover, we employ a sign-text alignment loss to further narrow the gap between the sign tokens and text tokens.
These designs enable SignLLM to produce sign sentences that embody two key characteristics of spoken languages: discrete characteristics and a hierarchical structure, which enhances their compatibility with LLMs and makes them more readily interpretable by LLMs.

After producing the language-like sign sentences, we feed them into an off-the-shelf and frozen LLM along with a text prompt that instructs the LLM to generate translations in the desired language.
We empirically observe that, by employing our SignLLM's designs to align sign videos with languages, we can already leverage a frozen LLM to attain state-of-the-art SLT performance.
These findings suggest that our proposed SignLLM framework is a promising first step towards effectively harnessing LLMs for SLT.
We hope our initial explorations can inspire future work within the community to leverage LLMs for SLT.

In summary, our main contributions are:
(1) We propose a novel SignLLM framework that is the first to harness the power of off-the-shelf and frozen LLMs for SLT.  
(2) To make the input sign video compatible with LLMs, our SignLLM framework incorporates two designs: a VQ-Sign module to quantize the sign video into a sequence of discrete character-level sign tokens and a CRA module that transforms the character-level sign tokens to word-level sign tokens.
(3) Through our proposed designs, we achieve state-of-the-art gloss-free results on two popular SLT datasets.

\section{Related Work}

\textbf{Sign Language Translation (SLT)} aims to transform sign videos into natural language sentences. 
It is a challenging task that requires understanding of both visual and linguistic cues \cite{SLT:glossfree,SLT:SignBT,SLT:SLTUNET}, and the challenge is exacerbated by the limited availability of paired sign-text data \cite{camgoz2020sign,SLT:SLTUNET,yin2020better,SLT:glossfree,SLT:MMTLB} which limits the performance of SLT methods.
To improve SLT performance, many previous works \cite{camgoz2018neural,ko2019neural,kan2022sign,camgoz2020sign,li2020tspnet,yin2020better,fu2023token,tang2021graph,SLT:TwoStream,SLT:SLTUNET,SLT:glossfree,SLT:STMC-T} aim to enhance the visual sign representations and text decoding capabilities of SLT methods. 
Some works propose deep architectures based on RNNs \cite{camgoz2018neural,ko2019neural}, GCNs \cite{kan2022sign}, and Transformers \cite{camgoz2020sign,li2020tspnet,yin2020better,voskou2021stochastic}.
Other approaches include introducing a keypoint estimator to enhance the visual sign representation \cite{ko2019neural,SLT:TwoStream,tang2021graph,SLT:STMC-T}, introducing pre-training tasks \cite{SLT:glossfree,fu2023token}, or jointly modelling several SLT-related tasks \cite{SLT:SLTUNET}.
{Some works also introduce larger datasets (e.g., How2Sign \cite{duarte2021how2sign} and BOBSL \cite{albanie2021bbc}), which present a huge challenge with their large sign and text vocabularies.}
Besides, some recent works \cite{yin2023gloss,SLT:glossfree} focus on the gloss-free setting -- these works do not use sign gloss annotations for training, which reduces the cost of training SLT models, and our work also falls into this category.
In contrast to existing works, we aim to harness the capabilities of off-the-shelf and frozen LLMs to perform SLT, by regularizing the sign videos into a language-like representation and prompting the LLM to generate text of the desired language.

\begin{figure*}[th]
  \centering
  \includegraphics[width=0.88\linewidth]{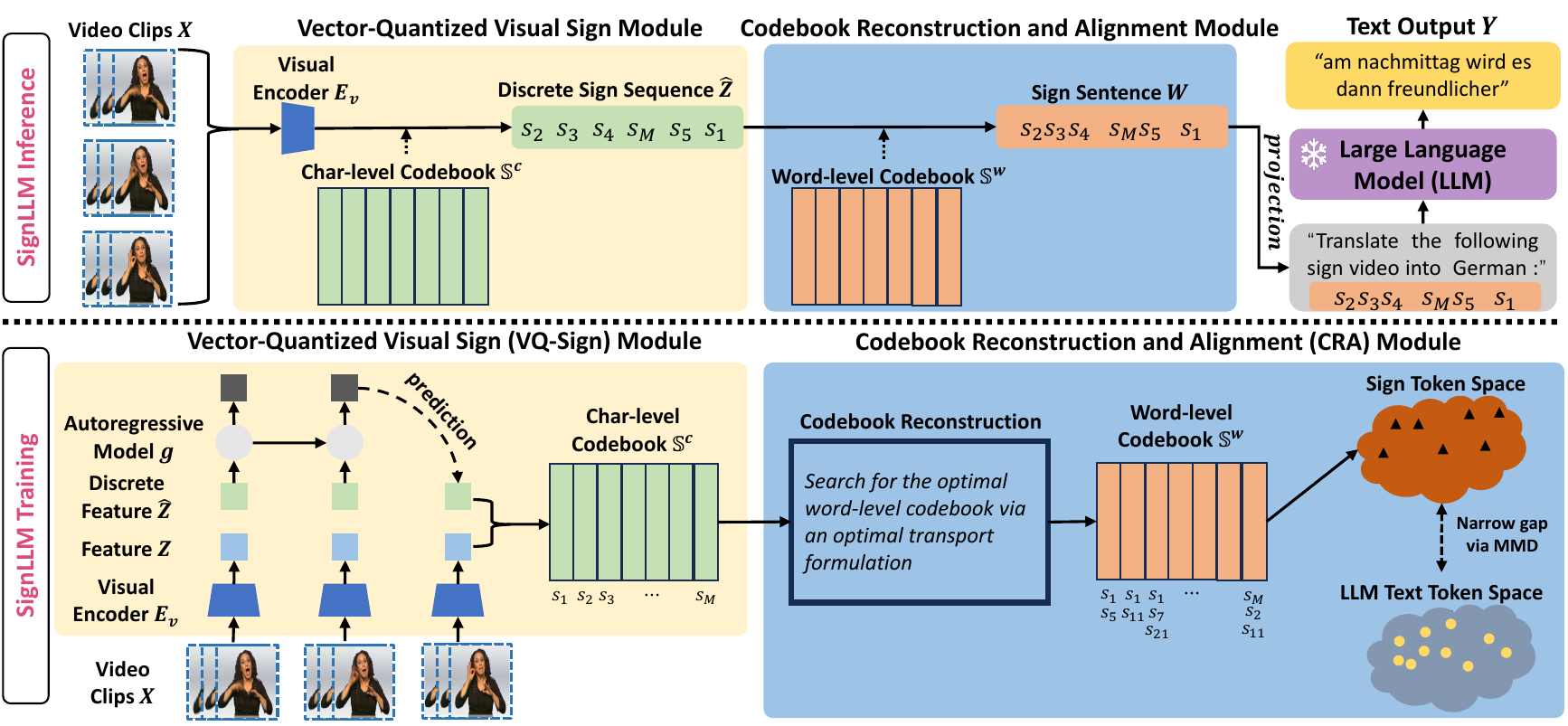}
  \vspace{-3mm}
  \caption{
  An overview of our SignLLM framework.
   \textbf{During inference (top):} Given an input sign video $X$, we first pass it through our VQ-Sign module to obtain a sequence of discrete character-level sign tokens $\hat{Z}$. Our VQ-Sign consists of a visual encoder $E_v$ to extract compact features and a character-level sign codebook $\mathbb{S}^c$ for quantization to obtain $\hat{Z}$. 
   Next, we feed $\hat{Z}$ into our CRA module, which reorganizes $\hat{Z}$ by replacing short sequences of character tokens with word-level tokens via the word-level codebook, e.g., character sequence $[s_2,s_3,s_4]$ to word $s_2s_3s_4$. 
   This transforms the sign video data to a language-like sign sentence $W$, which is fed into the LLM along with a text prompt which guides the LLM to generate translations in the desired language.      
   \textbf{During training (bottom):} We optimize VQ-Sign and its discrete sign codebook via a context prediction task, which seeks to recognize the future time steps based on the current context information.      
   Next, for our CRA module, we construct the optimal word-level codebook by considering two aspects: entropy and size, which we address using optimal transport techniques.
   Then, we narrow the gap between the sign token space and LLM's text token space via minimizing the MMD loss, which improves the semantic compatibility between them.
    }
    \vspace{-6mm}
    \label{fig:overall_pipeline}
\end{figure*}

\textbf{Large Language Models (LLMs)} refer to language models that have been trained extensively on a very large web-scale text corpus.
LLMs have shown impressive text generation capabilities, attracting a lot of attention recently \cite{zhao2023survey,foo2023ai,qu2024llmar}.
In particular, since LLMs have been trained on large amounts of text data, they demonstrate strong generalization capabilities across various text-based tasks, including code generation \cite{roziere2023code}, open-domain question answering \cite{zhu2021retrieving}, and multilingual translation 
\cite{chowdhery2022palm,brown2020language}.
Inspired by the recent advancements in LLMs, we explore harnessing LLMs for translation of sign videos, by converting the sign videos to a sequence of language-like sign tokens via our SignLLM framework, and treating the sign tokens as a form of language that can be translated by an LLM.
To the best of our knowledge, we are the first work to leverage an off-the-shelf and frozen LLM to tackle SLT.

\def\mA{{\bm{A}}}
\def\mB{{\bm{B}}}
\def\mC{{\bm{C}}}
\def\mD{{\bm{D}}}
\def\mE{{\bm{E}}}
\def\mF{{\bm{F}}}
\def\mG{{\bm{G}}}
\def\mH{{\bm{H}}}
\def\mI{{\bm{I}}}
\def\mJ{{\bm{J}}}
\def\mK{{\bm{K}}}
\def\mL{{\bm{L}}}
\def\mM{{\bm{M}}}
\def\mN{{\bm{N}}}
\def\mO{{\bm{O}}}
\def\mP{{\bm{P}}}
\def\mQ{{\bm{Q}}}
\def\mR{{\bm{R}}}
\def\mS{{\bm{S}}}
\def\mT{{\bm{T}}}
\def\mU{{\bm{U}}}
\def\mV{{\bm{V}}}
\def\mW{{\bm{W}}}
\def\mX{{\bm{X}}}
\def\mY{{\bm{Y}}}
\def\mZ{{\bm{Z}}}
\def\mBeta{{\bm{\beta}}}
\def\mPhi{{\bm{\Phi}}}
\def\mLambda{{\bm{\Lambda}}}
\def\mSigma{{\bm{\Sigma}}}

\section{Method}

In this section, we first introduce the overview of our SignLLM in Sec.~\ref{sec:SignLLM_pipeline}. 
Then, we describe two main components of our SignLLM framework: the VQ-Sign and CRA modules in Sec.~\ref{sec:VQ-Sign} and Sec.~\ref{sec:CRA}, respectively. 
Finally, we list the training and inference details in Sec.~\ref{sec:training_and_inference}.

\subsection{SignLLM Overview}
\label{sec:SignLLM_pipeline}

In order to effectively handle SLT, in this paper we draw inspiration from LLMs' remarkable capabilities in generating translations across multiple languages \cite{chowdhery2022palm,brown2020language}.
In particular, LLMs have been extensively trained on a large web-scale multilingual text corpus and have learned diverse knowledge regarding the properties of many languages, thus they are able to draw on shared commonalities with previously learned languages to effectively handle new languages with limited data \cite{yang2023bigtrans, zhu2023multilingual}.

Therefore, to leverage the strong translation capabilities of LLMs to handle SLT, we introduce a novel SignLLM framework.
SignLLM converts the input sign video $X$ into a language-like sign sentence $W$ that aligns with the linguistic characteristics of spoken languages and is friendly and compatible with LLMs.
Then, to perform SLT, the language-like sign sentence $W$ can be fed into an off-the-shelf and frozen LLM along with a text prompt that guides the LLM to generate translations in the desired language.

Specifically, to produce sign sentences $W$ that are friendly and understandable by LLMs, we aim to regularize our sign sentences $W$ to embody two core linguistic characteristics:
\textbf{Discrete Characteristics:} Spoken languages are naturally discrete and consist of distinct words or sub-words with corresponding discrete tokens in a vocabulary \cite{van2017neural,bostrom2020byte}. 
\textbf{Hierarchical Structure}: Most spoken languages exhibit three hierarchical semantic levels -- the sentence, word, and character levels \cite{levelt1999producing,seneff1998use}, where words are composed from characters and sentences are composed from various words.

To achieve the above, our SignLLM framework comprises of three parts, as shown in Fig.~\ref{fig:overall_pipeline}: 
(1) The \textbf{VQ-Sign module} converts the input sign video $X$ into a sequence of discrete sign tokens $\hat{Z}$, aligning the sign representations with text's \textit{discrete characteristics}. These sign tokens $\hat{Z}$ are character-level sign tokens that are retrieved from a learned discrete character-level codebook.
(2) The \textbf{CRA module} maps meaningful compositions of character-level sign tokens $\hat{Z}$ into word-level sign tokens that form a sign sentence $W$, further imparting a language-like \textit{hierarchical structure} to the video sign representations.
Moreover, we also align the sign token codebooks towards the text token space to improve semantic compatibility. 
(3) An \textbf{off-the-shelf LLM} takes the sign sentence $W$ as input, along with an instructive text prompt that guides the LLM to generate the translation in the desired language.
{More details regarding the text prompt are in the Supplementary.}
Next, we present our VQ-Sign and CRA modules in detail.

\subsection{Vector-Quantized Visual Sign Module}
\label{sec:VQ-Sign}

First, in order to produce a language-like representation, we would like to impart \textit{discrete characteristics} to the input sign videos, aligning them more closely with spoken language representations that are inherently discrete and consist of distinct tokens in a vocabulary.
However, achieving this is not straightforward because sign videos are a continuous signal in a high-dimensional spatio-temporal space which cannot be easily represented by a set of discrete tokens, and for which the vocabulary is not readily available.
Hence, we introduce our Vector-Quantized Visual Sign (VQ-Sign) module to quantize the sign video $X$ into a sequence of discrete sign tokens $\hat{Z}$ via a sign codebook $\mathbb{S}^c$.
As illustrated in Fig.~\ref{fig:overall_pipeline}, our VQ-Sign module involves a series of steps, which we present in detail next.

In the first step, we extract a compact feature $Z$ from the high-dimensional input sign video $X \in \mathbb{R}^{N \times H \times W}$, where $N$ is the number of video frames, while $H$ and $W$ are the height and width of the video frames respectively.
To be precise, the sign video $X$ is first organized into a sequence of short overlapping video clips, then each short video clip is fed into a visual encoder $E_v$ to extract a compact feature representation of dimensionality $d$. 
Overall, this step transforms the original high-dimensional input sign video $X \in \mathbb{R}^{N \times H \times W}$ into a compact feature $Z \in \mathbb{R}^{\frac{N}{n}\times d}$, where {$n$ represents the number of frames between the start of neighboring clips}.
Notably, since $Z$ is obtained by processing $\frac{N}{n}$ short clips, $Z$ can also be seen as a sequence of $\frac{N}{n}$ clip-wise features, i.e., $\{z_t\}_{t=1}^{\frac{N}{n}}$, where each $z_t \in \mathbb{R}^{d}$ corresponds to the feature of the $t$-th short clip.

In the next step, we transform the feature $Z = \{z_t\}_{t=1}^{\frac{N}{n}}$ into a sequence of discrete tokens $\hat{Z}$ using a codebook $\mathbb{S}^{c}$. 
Specifically, we discretize each clip's feature $z_t$ into a discrete token $\hat{z}_t$ by finding the matching token $s_j$ from the codebook $\mathbb{S}^{c} = \{ 
s_i \}_{i=1}^M$, where the $i$-th token in the codebook is denoted as $s_i \in \mathbb{R}^{d}$ and $M$ is the number of tokens in the codebook.
The matching token $s_j$ is the codebook's closest element to the feature $z_t$ in terms of Euclidean distance, i.e., $j = \mathop{\text{arg\,min}} _i ( \| z_t-s_i \|^2_2 )$.
After the matching, each feature $z_t$ is replaced by $\hat{z}_t = s_j$, which results in a discrete token sequence as shown in Fig.~\ref{fig:overall_pipeline}, e.g., $[s_2,s_3,s_4,s_M,s_5,s_1]$.
Note that, we randomly initialize all the tokens $\{ 
s_i \}_{i=1}^M$ in the sign codebook $\mathbb{S}^c$ at the start and optimize them during training, as introduced next.

However, we face a challenge in learning the discrete codebook $\mathbb{S}^{c}$.
In particular, although autoencoding \cite{van2017neural} has been a popular method for producing a codebook of discrete units, the high complexity of sign videos makes autoencoding (i.e., self-reconstruction of sign videos) challenging and costly.
Therefore, inspired from predictive coding \cite{atal1970adaptive}, a widely-used method in text and speech representation learning \cite{baevski2022data2vec, mikolov2013efficient, oord2018representation}, we propose to learn discrete representations of sign videos through a context prediction task. 
Context prediction \cite{oord2018representation} is a self-supervised task that focuses on recognizing the future content in latent space based on the current information, which can learn discrete representations while eliminating the need for reconstructing the high-dimensional input video data.
Furthermore, previous works show that training with context prediction effectively captures the temporal dependencies and relationships between elements in a sequence \cite{baevski2020wav2vec,hannun2017sequence}, and the learned representations are often transferable to downstream tasks \cite{oord2018representation,baevski2022data2vec}.

Specifically, we employ a context prediction task where we try to distinguish a future sample $z_{\tau+k}$ based on the current context representation $c_\tau$ at various time steps $\tau$.
To facilitate this task, after we obtain the discrete token sequence $\hat{Z}$, we further produce a context latent representation $c_\tau$ using an auto-regressive model $g$ that summarizes all the discrete tokens before a certain time step $\tau$ (i.e., $\{\hat{z}_t \}_{t \leq \tau }$) to produce context latent representations $c_{\tau} = g(\{\hat{z}_t \}_{t \leq \tau })$.
Then, we optimize our module by minimizing the following context prediction contrastive loss $\mathcal{L}_k^{\text{cp}}$:

\vspace{-5mm}
\setlength{\abovedisplayskip}{0pt}
\setlength{\belowdisplayskip}{0pt}
\begin{equation}
\small
\mathcal{L}_k^{\text{cp}} =  - \sum_{\tau =1}^{\frac{N}{n} - k}  \bigl(
\log \sigma  (z_{\tau +k}^{\top} h_{\tau}) + \lambda \mathop{\mathbb{E}}\limits_{\widetilde{z} \sim p_n}\ [ \log \sigma(-\widetilde{z}^\top h_{\tau}) ]\ \bigl)\,
\end{equation}
where $h_{\tau}$ is obtained by applying a trainable linear layer to $c_\tau$, $\sigma (z_{\tau +k}^\top h_\tau)$ is the probability of $z_{\tau +k}$ being the true sample among the negatives, 
$\widetilde{z}$ are negative samples drawn from a minibatch $p_n$, and $\lambda$ is a hyperparameter.
We sum $\mathcal{L}_k^{\text{cp}}$ over different step sizes $k$ to obtain the context prediction loss $\mathcal{L}^{\text{cp}} = \sum_{k=1}^{K} \mathcal{L}_k^{\text{cp}}$, where $K$ is the maximum number of future clips that we are interested in.

Following \cite{devlin2019bert}, in order to optimize the matching between $\hat{z}_t$ and $z_t$, we further add two losses to optimize the matching distance between $\hat{z}_t$ and $z_t$, such that the overall loss $\mathcal{L}^{VQ}$ to optimize our VQ-Sign module is as follows:

\vspace{-5mm}
\setlength{\abovedisplayskip}{0pt}
\setlength{\belowdisplayskip}{0pt}
\begin{equation}
\small
\mathcal{L}^{VQ} = \mathcal{L}^{\text{cp}} + \sum_{t=1}^{N/n}  \| \text{sg}(z_t) - \hat{z}_t \|^2
    + \gamma \sum_{t=1}^{N/n}  \| z_t - \text{sg}(\hat{z}_t) \|^2,
\label{eq:vq_loss}
\end{equation}
where $\text{sg}(z) \equiv z$ is the stop-gradient operator and $\gamma$ is a hyperparameter.
By optimizing $\mathcal{L}^{VQ}$, we can train our VQ-Sign and the discrete codebook without the need for reconstructing high-dimensional video clips, which makes the codebook construction a viable and relatively cheap option.

In summary, our VQ-Sign transforms sign videos into sequences of discrete sign tokens $\hat{Z}$, which are friendlier and more understandable to LLMs.
Notably, the \textit{produced discrete tokens $\hat{Z}$ can be likened to character-level tokens}, in the sense that each discrete token $\hat{z}_t$ corresponds to a short clip and may not contain much semantic meaning on its own (similar to linguistic characters), but they can be combined into a sequence to convey a clear semantic meaning (akin to forming a word or sentence).
Thus, inspired by this, we call VQ-Sign's codebook $\mathbb{S}^c$ the character-level codebook that contains character-level sign tokens.

\subsection{Codebook Reconstruction and Alignment}
\label{sec:CRA}

In the previous section, we quantize sign videos into discrete character-level sign tokens, which aligns them closer to language representations. 
In this section, our goal is to impart a \textit{hierarchical structure} to our sign video representations, which makes them align even closer to language representations.
Specifically, we aim to compose our character-level sign tokens into word-level sign tokens to mirror the observed hierarchical structure in spoken language, which makes them even more compatible with LLMs.

Intuitively, considering a spoken language sentence, we can represent it as a sequence of words, with each word formed by one or multiple characters. 
For example, a sentence `I love AI' can be decomposed into a word sequence $[\text{`I',`love',`AI'}]$, where the word `love' is in turn formed by the character sequence $[\text{`l',`o',`v',`e'}]$.
We observe that, although each individual character may not contain much semantic meaning on its own, they can be composed to form words with clearer semantic meaning.
In a similar fashion, we also want to impart such hierarchical structure to our character-level sign tokens by composing them to form meaningful word-level sign tokens.

Hence, we aim to find an optimal transformation from character-level sign tokens to word-level sign tokens for enhanced readability and compatibility with LLMs.
To this end, we introduce the Codebook Reconstruction and Alignment (CRA) module to transform the character-level codebook $\mathbb{S}^{c}$ from VQ-Sign into a word-level codebook $\mathbb{S}^{w}$ whose tokens convey richer and clearer semantic meaning.
Inspired by optimal transport methods \cite{cuturi2013sinkhorn,villani2009optimal,xu2021vocabulary}, we observe that the above transformation can be formulated as an optimal transport problem of transporting characters into words, thus we introduce a \textit{codebook reconstruction algorithm} with an optimal transport formulation to find an optimal transformation.
Additionally, to further reduce the distribution gap between the sign tokens and the text tokens, our CRA module also performs \textit{sign-text alignment}, enhancing the semantic compatibility of the sign tokens with LLMs.
We introduce the details below.

To begin, the objective of our \textbf{Codebook Reconstruction Algorithm} is to create a word-level codebook $\mathbb{S}^{w}$ based on VQ-Sign's character-level codebook $\mathbb{S}^{c}$. 
The challenge lies in determining which character-level sign tokens should be assembled together to form word-level sign tokens, which is a complex problem.
To address this complexity, we adopt an approach based on two fundamental principles.
Firstly, in order to maximize the overall predictability of the word-level tokens and enhance the distinctiveness of each token, we seek to minimize the \textit{entropy} of each word-level token within the vocabulary \cite{martin2011mathematical}.
We remark that, several approaches for establishing language-based subword vocabularies \cite{bostrom2020byte,song2021fast} can be seen as entropy-minimizing approaches, employing different heuristics to establish the vocabulary with the goal of minimizing entropy \cite{gage1994new}.  
On the other hand, considering the limited availability of sign video data, we incorporate \textit{codebook size} as another key factor in our word-level codebook construction, since studies on languages with limited data \cite{sennrich2019revisiting, haddow2022survey} have also identified vocabulary size as a crucial aspect. 
Specifically, too small a vocabulary can result in sub-optimal entropy values, while an excessively large vocabulary size can lead to issues such as parameter explosion and token sparsity which hinder understanding \cite{allison2006another}, and finding the right balance between these effects becomes even more sensitive for languages with limited data \cite{ding2019call,sennrich2019revisiting}.

Based on these principles, our objective is to determine an optimal codebook size that maximizes the entropy reduction while taking into account the increase in codebook size.
In other words, we would like to find an optimal codebook size that maximizes the gradient of the entropy reduction with respect to the codebook size increase.
To simplify the optimal size searching problem, we define a fixed size increment $m$ and search through codebooks of various sizes (where the difference between each codebook size is $m$ tokens).
Specifically, we define the $r$-th codebook ($\mathbb{S}^{w}_r$) as the codebook with $r \times m$ tokens.
Then, we seek to identify the optimal set of word-level tokens, where each word-level token is composed of character-level tokens.
We approach this by formulating the character compositions as an optimal transport problem, where characters are transported to words.

However, it can be challenging to identify specific character combinations that convey precise semantic information, due to the temporal complexity of sign videos, which often makes the character-level token sequences $\hat{Z}$ quite messy. 
For instance, some signers may execute the signing motions at a slower speed, which can lead to consecutive short video clips being highly similar, resulting in consecutively-repeating character-level discrete tokens.
Thus, the character-level sequences between different signers can differ significantly {(e.g., $[s_1,s_2]$ vs $[s_1, s_1, s_1,s_2,s_2]$)} due to such duplication of character-level tokens, even though they may contain the same semantic information.
At the same time, simply filtering out repeated character-level tokens straightforwardly {(e.g., setting all $[s_1, s_1, s_1,s_2,s_2]$ to $[s_1,s_2]$)}  is sub-optimal, since the speed of some signs can also convey some information  \cite{hou2019signspeaker,vicars2017asl}, e.g., if a signer signs ``ugly'' quickly, it conveys ``very ugly'' in American Sign Language.

\noindent
\textbf{Pre-processing of Repeated Characters.}
Therefore, to alleviate the impact of the signer's speed while keeping the information regarding each sign's speed, we \textit{first pre-process the character-level sequences} as follows: 
First, we find all the repeated tokens in the character-level sequence and compute the average number of repeated tokens ($\alpha$) in each sequence. 
Then, for each repeated sequence (e.g., $[s_1, s_1, s_1]$), we keep the first character and remove the tailing repetitive ones (e.g., $[s_1, s_1, s_1]$ to $[s_1]$).
At the same time, if the character-level tokens repeat more than $\alpha$ times, we insert a single character-level token $s_0$ as a ``slowing down'' sign, e.g., $[s_1, s_1, s_1]$ to $[s_1,s_0]$ if $\alpha < 3$.
Crucially, this allows us to reduce redundancy, while still representing ``quick'' or ``slow'' signs that account for differences in the signer's speed.
Overall, this pre-processing and reducing of repeating characters makes the character-level sequence less messy, facilitating the search for specific meaningful character combinations.

\noindent
\textbf{Optimal Transport Formulation.}
Then, using the pre-processed character-level sequences, we search for an optimal word-level codebook via an optimal transport formulation \cite{cuturi2013sinkhorn,villani2009optimal,xu2021vocabulary}.
Following our discussion above, we aim to find an optimal word-level codebook (i.e., $\mathbb{S}^w_r$) with low entropy and compact size.
Specifically, to measure codebook entropy,
we follow \cite{gutierrez2021characters,martin2011mathematical,nag2023entropy} to define the entropy of the $r$-th codebook $\mathbb{S}^{w}_r$ as:

\vspace{-2mm}
\setlength{\abovedisplayskip}{0pt}
\setlength{\belowdisplayskip}{0pt}
\begin{equation}
\small
\mathcal{H}_{\mathbb{S}^{w}_r} = - \sum_{w_j\in \mathbb{S}^{w}_r} P(w_j)\log P(w_j),
\label{eq:entropy}
\end{equation}
where $P(w_j)$ is the relative frequency of the $j^{th}$ token $w_j$ from the word-level codebook $\mathbb{S}^{w}_r$. 
Then, based on VQ-Sign's character-level codebook $\mathbb{S}^{c}$, the entropy of the word-level codebook $\mathbb{S}^{w}_r$ can be computed through the following (with proof in Supplementary):

\vspace{-5mm}
\setlength{\abovedisplayskip}{0pt}
\setlength{\belowdisplayskip}{0pt}
\begin{equation}
\small
\begin{split}
\mathcal{H}_{\mathbb{S}^{w}_r}
=& - {\sum_{w_j\in \mathbb{S}^{w}_r} \sum_{s_i\in \mathbb{S}^{c}} P(w_j,s_i)\log P(w_j, s_i)} \\
& - { \sum_{w_j\in \mathbb{S}^{w}_r} \sum_{s_i\in \mathbb{S}^{c}}P(w_j,s_i)(-\log P(s_i | w_j))} ,\\
\end{split}
\label{eq:decomposition}
\end{equation}
where $P(s_i | w_j)$ is the probability of the character-level token $s_i$ appearing in the word-level token $w_j$.

Then, to formally define our objective function, we follow existing works \cite{peyre2019computational,cuturi2013sinkhorn,xu2021vocabulary} to define a transport matrix $\mP$ that represents the assignments of characters to words, and a distance matrix $\mD$ that represents the cost of transportation.
We define the transport matrix $\mP \in \mathbb{R}^{m \times (r \cdot m)}$ with the $(j, i)$-th element as $P(w_j,s_i)$, and define the distance matrix $\mD \in \mathbb{R}^{m \times (r \cdot m)}$ as a matrix whose $(j, i)$-th element is $\log P(s_i | w_j)$.
Note that, if $w_j$ contains $s_i$, we use $\frac{1}{\text{length}(w_j)}$ to estimate $P(s_i | w_j)$, and if $w_j$ does not contain $s_i$, then it is deemed an infeasible assignment, and we set the distance $\log P(s_i | w_j) = \infty$.
We further define $H(\mP)$ as $- {\sum_{w_j\in \mathbb{S}^w_r} \sum_{s_i\in \mathbb{S}^c} P(w_j,s_i)\log P(w_j, s_i)}$, which is simply the entropy of the probability distribution $P(w_j,s_i)$.

Hence, based on Eq.~\ref{eq:decomposition}, the objective function for minimizing the entropy of $\mathbb{S}^{w}_r$ can be formulated as:

\vspace{-2mm}
\setlength{\abovedisplayskip}{0pt}
\setlength{\belowdisplayskip}{0pt}
\begin{equation}
\small
\mathop{\text{arg\,min}}\limits_{\mP \in \mathbb{R}^{m \times (r \cdot m)}} H(\mP) + \sum_{j}\sum_{i}\mP(j,i)\mD(j,i).
\label{eq:objective_function_2}
\end{equation}
Specifically, following previous works \cite{peyre2019computational,cuturi2013sinkhorn,xu2021vocabulary}, we impose two constraints on the transport matrix $\mP$: the sum of each row in $\mP$ should equal to the probability of character token $s_i$ and the sum of each column in $\mP$ should equal to the probability of word token $w_j$. Formally, we constrain the transport matrix $\mP$ with: $|\sum _i \mP(i, j) - P(w_j)| \leq \epsilon $ and $ |\sum _j \mP(i, j) - P(s_i)| \leq \epsilon,$ where $\epsilon$ is a small positive fixed hyperparameter.

Intuitively, this optimization process can be regarded as an optimal transport problem to ﬁnd the best way to transport mass from $\mathbb{S}^{c}$ to $\mathbb{S}^{w}_r$.
To handle this optimal transport problem, we leverage the Sinkhorn algorithm \cite{cuturi2013sinkhorn,peyre2019computational,xu2021vocabulary}, allowing us to effectively construct the candidate word-level codebooks $\mathbb{S}^{w}_{r}$ with minimal entropy.
Since our increment $m$ between each candidate codebook is fixed, we can find the optimal codebook size that maximizes the gradient of entropy reduction by simply computing and finding the maximum the entropy difference between $\mathbb{S}^{w}_{r}$ and $\mathbb{S}^{w}_{r-1}$.

After finding the optimal word-level codebook, we construct the word-level sign tokens by composing all features of the character-level tokens into the word-level tokens via our autoregressive model $g$. 
Refer to Supplementary for more details. 
Overall, with our codebook reconstruction algorithm, we can construct an optimal word-level sign codebook with low entropy yet also with relatively small size.

\vspace{2mm}
\noindent \textbf{Sign-Text Alignment.}
Next, we further align the sign tokens with the text tokens used in LLMs in order to further improve semantic compatibility between them. 
To achieve this, we measure the distribution gap between sign tokens and text tokens via Maximum Mean Discrepancy (MMD) \cite{tolstikhin2016minimax} and then optimize the sign tokens' embeddings by minimizing MMD, which narrows down the distribution gap. 
Specifically, we compute the gap between the sign embedding space $\mathbb{F}^s$ and text embedding space $\mathbb{F}^t$ via MMD as:

\vspace{-5mm}
\setlength{\abovedisplayskip}{0pt}
\setlength{\belowdisplayskip}{0pt}
\begin{equation}
\label{eq:mmd_loss}
\small
\begin{split}
\mathcal{L}^{MMD}&(\mathbb{F}^s,\mathbb{F}^t)
= \sum_{i=1}^{n_s}\sum_{j=1}^{n_s} \frac{k(f(p_i),f(p_j))}{n_s^2} \\
+&  \sum_{i=1}^{n_t}\sum_{j=1}^{n_t} \frac{k(q_i,q_j)}{n_t^2} - \sum_{i=1}^{n_s}\sum_{j=1}^{n_t} \frac{2 \cdot k(f(p_i),q_j)}{n_s n_t},\\
\end{split}
\end{equation}
where $p$ and $q$ are the tokens in $\mathbb{F}^s$ and $\mathbb{F}^t$, $f$ is a small projection module that projects $\mathbb{F}^s$ to $\mathbb{F}^t$,
$n_s$ and $n_t$ are the numbers of tokens in $\mathbb{F}^s$ and $\mathbb{F}^t$, and $k(\cdot)$ represents the radial kernel~\cite{tolstikhin2016minimax} that measures the distance between two samples. 
We apply MMD loss to both the word-level and the character-level tokens to narrow the overall sign-text gap.

\subsection{Training and Inference}
\label{sec:training_and_inference}

\noindent \textbf{Inference.} Given a sign video $X$, we first extract compact features $Z$ via the visual encoder $E_v$, and quantize $Z$ to $\hat{Z}$ via VQ-Sign's learned character-level codebook. 
Then, we transform the sequence of discrete character-level tokens $\hat{Z}$ into word-level tokens via our CRA, which produces a sign sentence $W$.
Lastly, we project the sign sentence $W$ to LLM embedding space via the small projection module $f$, and then feed the sign sentence $W$ into the LLM along with a text prompt to instruct the LLM to perform the SLT task and generate text in the desired language.

\noindent \textbf{Training.} 
Our SignLLM is optimized in two stages: pre-training and fine-tuning.
Specifically, the pre-training stage, which does not require explicit SLT supervision, includes two sub-stages:
(i) We first pre-train VQ-Sign via the context prediction task with $\mathcal{L}^{VQ}$ in Eq.~\ref{eq:vq_loss}.
(ii) Then, based on the VQ-Sign's learned character-level codebook, we construct the word-level codebook with the codebook reconstruction algorithm (Sec.~\ref{sec:CRA}) and apply the MMD loss $\mathcal{L}^{MMD}$ in Eq.~\ref{eq:mmd_loss} to align the sign codebooks ($\mathbb{S}^{c}$ and $\mathbb{S}^{w}$) and the text vocabulary of the desired language.

After the pre-training, we fine-tune SignLLM.
To aid LLMs in understanding the sign sentences as texts, we additionally maximize the similarity between the text tokens $Y$ generated by LLM and the ground truth tokens $\bar{Y}$ as: $\mathcal{L}^{sim}$ by minimizing the cross-entropy between them.
We fine-tune our SignLLM (with frozen LLM) via the loss $\mathcal{L}^{ft}$ as follows: $\mathcal{L}^{ft} = \mathcal{L}^{VQ} + \lambda _1 \mathcal{L}^{MMD} + \lambda _2 \mathcal{L}^{sim}$,
where $\lambda_1$ and $\lambda_2$ are hyperparameters.
Note that we follow previous gloss-free works \cite{yin2023gloss,SLT:glossfree} to train our SignLLM framework, eliminating the need for additional gloss data.

\section{Experiments}
\subsection{Implementation Details}

Our visual encoder $E_v$ is constructed by appending two Conv3D layers with a kernel size of (5, 3, 3) and a stride of (2, 1, 1) to a ResNet18 \cite{he2016deep} pre-trained on ImageNet \cite{deng2009imagenet}. Each clip consists of 13 frames and the gap between the neighboring clips ($n$) is $4$.
Besides, the auto-regressive model $g$ is implemented as a Convolutional Gated Recurrent Layer with a kernel size of (1, 1). 
We set the total number of discrete vectors $M$ at 256 and each vector's dimension $d$ at 1024 for our character-level codebook.
In VQ-Sign's pre-training phase, we set $\gamma = 0.25$, initialize the learning rate at 0.01 and use the Adam algorithm, training the model to predict the future three clips ($K=3$) for 200 epochs. 
During codebook reconstruction, we set the increment $m$ to 32. 
We employ the frozen LLaMA-7B-16bit \cite{touvron2023llama} as our LLM and project the codebook space to LLaMA's embedding space via $f$, which consists of two fc-layers with ReLU. 
We set $\lambda_1=0.5$, $\lambda_2=1$, and initialize the learning rate at 0.001 to fine-tune our SignLLM over 20 epochs. 
Please see Supplementary for more implementation details.

\begin{table*}[h]
\centering
\scriptsize
\caption{Results on Phoenix-2014T dataset \cite{camgoz2018neural}.}
\vspace{-4mm}
\resizebox{0.75\textwidth}{!}{
\begin{tabular}{l|l|ccccc|ccccc}
\hline
\multirow{2}{*}{Setting} & \multirow{2}{*}{Method} & \multicolumn{5}{c|}{Dev} &  \multicolumn{5}{c}{Test} \\
\cline{3-12}
 &  & B1  & B2 & B3 & B4 & ROUGE  & B1  & B2 & B3 & B4 & ROUGE   \\
\hline
\multirow{2}{*}{Gloss-based} &  SLRT \cite{camgoz2020sign} & 47.26  &  34.40 &  27.05 &  22.38 &  -  & 46.61 &  33.73  & 26.19 &  21.32  &  -   \\
  & ConSLT \cite{fu2023token}  & - & - & - & 21.11 & 47.74 & - & - & - & 21.59 & 47.69 \\
 & STN-SLT \cite{voskou2021stochastic} &  49.12  & 36.29  &  28.34 &  23.23 &  - &  48.61 &  35.97 &  28.37 &  23.65 &  - \\
 &  STMC-T \cite{SLT:STMC-T} &  47.60 &  36.43 &  29.18 &  24.09 &  48.24 &  46.98 &  36.09 &  28.70 &  23.65 &  46.65  \\
 & BN-TIN-Transf.+SignBT \cite{SLT:SignBT}  &  51.11 &  37.90 &  29.80 &  24.45 &  50.29 &  50.80 &  37.75 &  29.72 &  24.32 &  49.54  \\
  & PET \cite{SLT:PET} & - & - & - & - & - & 49.54 & 37.19 & 29.30 & 24.02 & 49.97 \\
 & MMTLB \cite{SLT:MMTLB} &  53.95 &  41.12 &  33.14 &  27.61 &  53.10 &  53.97 &  41.75 &  33.84 &  28.39 &  52.65  \\
 &  TS-SLT \cite{SLT:TwoStream} &  54.32 &  41.99 &  34.15 &  28.66 &  54.08 &  54.90 &  42.43 &  34.46 &  28.95 &  53.48  \\
  & SLTUNET \cite{SLT:SLTUNET} & - & - & - & 27.87 & 52.23 & 52.92 & 41.76 & 33.99 & 28.47 & 52.11 \\
\hline
\multirow{2}{*}{Gloss-free} & NSLT \cite{camgoz2018neural} &  28.10 &  16.81 &  11.82 &  9.12 &  31.00 &  27.10 &  15.61 &  10.82 &  8.35 &  29.70 \\
 & NSLT+Bahdanau \cite{camgoz2018neural,bahdanau2015neural} &  31.87 &  19.11 &  13.16 &  9.94 &  31.80 &  32.24 &  19.03 &  12.83 &  9.58 &  31.80 \\
 & NSLT+Luong \cite{camgoz2018neural,luong2015effective} &  31.58 &  18.98 &  13.22 &  10.00 &  32.60 &  29.86 &  17.52 &  11.96 &  9.00 &  30.70 \\
 & TSPNet \cite{li2020tspnet} &  - &  - &  - &  - &  - &  36.10 &  23.12 &  16.88 &  13.41 &  34.96 \\
 & CSGCR \cite{zhao2021conditional} &  35.85 &  24.77 &  18.65 &  15.08 &  38.96 &  36.71 &  25.40 &  18.86 &  15.18 &  38.85 \\
 & GASLT \cite{yin2023gloss} &  - &  - &  - &  - &  - &  39.07 &  26.74 &  21.86 &  15.74 &  39.86 \\
 & GFSLT-VLP \cite{SLT:glossfree} &  44.08 &  33.56 &  26.74 &  22.12 &  43.72 &  43.71 &  33.18 &  26.11 &  21.44 &  42.49 \\
\cline{2-12}
& Ours &\textbf{46.88}  &\textbf{36.59}  &\textbf{29.91}  &\textbf{25.25}  &\textbf{47.23}  & \textbf{45.21} & \textbf{34.78} &\textbf{28.05} & \textbf{23.40} & \textbf{44.49}  \\ 
 \hline
\end{tabular}
}
\vspace{-4mm}
\label{table:main_results_phoenix}
\end{table*}

\begin{table*}[h]
\centering
\scriptsize
\caption{Results on CSL-Daily dataset \cite{SLT:SignBT}. * means that the result was reproduced by \cite{SLT:glossfree}}
\vspace{-4mm}
\resizebox{0.75\textwidth}{!}{
\begin{tabular}{l|l|ccccc|ccccc}
\hline
\multirow{2}{*}{Setting} & \multirow{2}{*}{Method} & \multicolumn{5}{c|}{Dev} &  \multicolumn{5}{c}{Test} \\
\cline{3-12}
 &  & B1  & B2 & B3 & B4 & ROUGE  & B1  & B2 & B3 & B4 & ROUGE   \\
\hline
\multirow{2}{*}{Gloss-based} &  SLRT \cite{camgoz2020sign} & 37.47 &  24.67 &  16.86 &  11.88 &  37.96 &  37.38 &  24.36 &  16.55 &  11.79 &  36.74   \\
  & ConSLT \cite{fu2023token}  & - & - & - & 14.80 & 41.46 & - & - & - & 14.53 & 40.98 \\
 & BN-TIN-Transf.+SignBT \cite{SLT:SignBT}  &  51.46 &  37.23 &  27.51 &  20.80 &  49.49 &  51.42 &  37.26 &  27.76 &  21.34 &  49.31  \\
 & MMTLB \cite{SLT:MMTLB} &  53.81 &  40.84 &  31.29 &  24.42 &  53.38 &  53.31 &  40.41 &  30.87 &  23.92 &  53.25  \\
 &  TS-SLT \cite{SLT:TwoStream} &  55.21 &  42.31 &  32.71 &  25.76 &  55.10 &  55.44 &  42.59 &  32.87 &  25.79 &  55.72  \\
 & SLTUNET \cite{SLT:SLTUNET} & - & - & - & 23.99 & 53.58 & 54.98 & 41.44 & 31.84 & 25.01 & 54.08   \\
\hline
\multirow{2}{*}{Gloss-free} & SLRT* \cite{camgoz2020sign} &  21.03 &  9.97 &  5.96 &  4.04 &  20.51 &  20.00 &  9.11 &  4.93 &  3.03 &  19.67 \\
 & NSLT+Luong \cite{camgoz2018neural,luong2015effective} &  34.22 &  19.72 &  12.24 &  7.96 &  34.28 &  34.16 &  19.57 &  11.84 &  7.56 &  34.54  \\
 & GASLT \cite{yin2023gloss} & - &  - &  - &  - &  - &  19.90 &  9.94 &  5.98 &  4.07 &  20.35 \\
 & GFSLT-VLP \cite{SLT:glossfree} &  39.20 &  25.02 &  16.35 &  11.07 &  36.70 &  39.37 &  24.93 &  16.26 &  11.00 &  36.44  \\
\cline{2-12}
 & Ours &  \textbf{42.45}  & \textbf{26.88} & \textbf{17.90} & \textbf{12.23} & \textbf{39.18} & \textbf{39.55} & \textbf{28.13} & \textbf{20.07} & \textbf{15.75} & \textbf{39.91} \\
 \hline
\end{tabular}
}
\vspace{-5mm}
\label{table:main_results_csldaily}
\end{table*}

\subsection{Datasets and Evaluation Metrics}

\noindent
\textbf{Datasets.} We follow previous works \cite{SLT:glossfree,SLT:TwoStream,SLT:STMC-T,SLT:SignBT,SLT:MMTLB,kan2022sign} to run experiments on the Phoenix-2014T \cite{camgoz2018neural} and CSL-Daily \cite{SLT:SignBT} datasets for SLT, and evaluate on their dev and test sets.
\textbf{Phoenix-2014T} \cite{camgoz2018neural} is a German sign language dataset with a vocabulary size of 2887 German words. The training, dev, and test sets contain 7096, 519, and 642 samples.
\textbf{CSL-Daily} \cite{SLT:SignBT} is a Chinese sign language dataset with a vocabulary size of 2343 Chinese words. The training, dev, and test sets contain 18401, 1077, and 1176 samples.

\noindent
\textbf{Evaluation Metrics.}
Following previous works \cite{SLT:glossfree,SLT:TwoStream,SLT:STMC-T,SLT:SignBT,SLT:MMTLB,li2020tspnet,yin2023gloss}, we adopt BLEU \cite{papineni2002bleu} and ROUGE-L \cite{lin2004rouge} as the evaluation metrics for SLT. 
BLEU-n evaluates the average translation precision up to n-grams, and we follow previous works \cite{SLT:glossfree,SLT:TwoStream,SLT:STMC-T,SLT:SignBT,SLT:MMTLB} to report results for BLEU-1 to BLEU-4 (i.e., B1, B2, B3 and B4).
ROUGE-L (or ROUGE) computes the F1 score based on the longest common subsequence between the predicted and ground truth texts.

\subsection{Main Results}

\noindent \textbf{Results on PHOENIX2014T dataset.} 
Tab.~\ref{table:main_results_phoenix} presents a comparison of our approach with state-of-the-art gloss-based and gloss-free methods for SLT. 
Our method consistently improves upon all reported metrics as compared to other gloss-free approaches.

\noindent \textbf{Results on CSL-Daily dataset.} 
We compare our method with state-of-the-art approaches on the CSLDaily dataset in Tab.~\ref{table:main_results_csldaily}. 
We outperform previous gloss-free works on all metrics, showing the efficacy of our approach.

\subsection{Ablation Study}

To further investigate the proposed method, we follow previous works \cite{SLT:SLTUNET,SLT:glossfree,SLT:PET,yin2023gloss,SLT:MMTLB} to conduct extensive ablation experiments on the Phoenix-2014T dev and test sets.
Refer to Supplementary for more experiment results.

\noindent\textbf{Impact of LLM.} First, we evaluate the impact of leveraging LLMs for the SLT task. Specifically, we establish three baselines: 
1) \textbf{Ours (w/o LLM)} where we replace the LLM with a trainable lightweight text generator (mBART \cite{liu2020multilingual}). 
2) \textbf{Ours (w/ T5)} where we replace our LLM with a smaller LLM (T5 \cite{raffel2020exploring}). 
As shown in Tab.~\ref{table:ab_LLMs}, our approach (w/ LLaMA \cite{touvron2023llama}) achieves a much better performance than Ours (w/o LLM), showing the efficacy of leveraging a powerful LLM.
We also find that using a smaller and less powerful LLM (T5) leads to worse performance, as expected.

\begin{table}[t]
\centering
\scriptsize
\caption{Ablation study for impact of LLM}
\vspace{-3mm}
\resizebox{0.483\textwidth}{!}{
\begin{tabular}{l|cccc|cccc}
\hline
\multirow{2}{*}{Method} & \multicolumn{4}{c|}{Dev} &  \multicolumn{4}{c}{Test} \\
\cline{2-9}
& B1  & B2 & B3 & B4  & B1  & B2 & B3 & B4  \\
\hline
Ours (w/o LLM) & 29.75  & 20.04 & 14.96 & 11.95 & 27.20 & 18.29 & 13.32 & 10.36  \\ 
Ours (w/ T5) & 44.20 & 34.55 & 27.15 & 22.90 & 44.03 & 34.12 & 27.23 & 22.51 \\ 
\hline
Ours & 46.88 & 36.59 & 29.91 & 25.25 & 45.21 & 34.78 & 28.05 & 23.40 \\ 
\hline
\end{tabular}
}
\label{table:ab_LLMs}
\vspace{-5mm}
\end{table}
\begin{table}[t]
\centering
\scriptsize
\caption{Ablation study for impact of SignLLM.}
\vspace{-3mm}
\resizebox{0.483\textwidth}{!}{
\begin{tabular}{l|cccc|cccc}
\hline
\multirow{2}{*}{Method} & \multicolumn{4}{c|}{Dev} &  \multicolumn{4}{c}{Test} \\
\cline{2-9}
& B1  & B2 & B3 & B4  & B1  & B2 & B3 & B4  \\
\hline
Encoder Only & 26.95  & 17.04 & 12.11 & 9.41 & 25.63 & 16.10 & 11.20 & 8.42 \\ 
Encoder + FT & 39.29  & 29.29  & 22.55 & 18.18 & 39.92 & 29.14 & 22.54 & 18.17  \\ 
 \hline
 Ours & 46.88 & 36.59 & 29.91 & 25.25 & 45.21 & 34.78 & 28.05 & 23.40   \\ 
 \hline
\end{tabular}
}
\label{table:ab_SignLLM}
\vspace{-5mm}
\end{table}
\begin{table}[t]
\centering
\scriptsize
\caption{Ablation study for main components of SignLLM.}
\vspace{-3mm}
\resizebox{0.483\textwidth}{!}{
\begin{tabular}{l|cccc|cccc}
\hline
\multirow{2}{*}{Method} & \multicolumn{4}{c|}{Dev} &  \multicolumn{4}{c}{Test} \\
\cline{2-9}
& B1  & B2 & B3 & B4  & B1  & B2 & B3 & B4  \\
\hline
Ours (w/o VQ-Sign) & 35.78 & 24.65 & 18.08 & 14.15 & 33.14 & 22.80 & 16.74 & 13.13 \\ 
Ours (w/o Codebook Reconstruction) & 40.45 & 30.40 & 24.07 & 19.79 & 40.25 & 30.05 & 23.63 & 19.47 \\ 
Ours (w/o Sign-text Alignment) & 29.05  & 19.33 & 13.72 & 10.40 & 28.67 & 19.22 & 13.73 & 10.63 \\ 
 \hline
 Ours & 46.88 & 36.59 & 29.91 & 25.25 & 45.21 & 34.78 & 28.05 & 23.40   \\  
 \hline
\end{tabular}
}
\label{table:ab_main_components}
\vspace{-7mm}
\end{table}

\noindent\textbf{Impact of SignLLM.} 
Next, we explore the impact of SignLLM by comparing against the following baselines:
1) \textbf{Encoder Only} where we directly feed the outputs of the visual encoder $E_v$ into the LLM, and keep the LLM frozen.
2) \textbf{Encoder + FT} where we directly feed the outputs of the visual encoder $E_v$ into the LLM, and follow LLaVA \cite{liu2024visual} to fine-tune them to translate sign videos via LoRA \cite{hu2022lora}.
We report the results in Tab.~\ref{table:ab_SignLLM}, where we significantly outperform the baselines. 
This shows that SignLLM is effective in harnessing off-the-shelf LLMs for the SLT task.

\noindent\textbf{Impact of Main Components of SignLLM.} 
We also verify the impact of the key components of SignLLM by comparing against the following baselines:
1) \textbf{Ours (w/o VQ-Sign)} where we directly quantize the visual encoder $E_v$'s output feature using $k$-means algorithm. The visual encoder $E_v$ is pre-trained via the similarity loss $\mathcal{L}^{sim}$ instead of our proposed VQ-Sign's loss $\mathcal{L^{VQ}}$. 
2) \textbf{Ours (w/o Codebook Reconstruction)} where we feed the character-level sign tokens $\hat{Z}$ from VQ-Sign directly into the LLM, while applying the sign-text alignment loss $\mathcal{L}^{MMD}$ to the character-level sign tokens.
3) \textbf{Ours (w/o Sign-text Alignment)} where we do not apply the sign-text alignment loss $\mathcal{L}^{MMD}$.
As shown in Tab.~\ref{table:ab_main_components}, removing any of our main designs leads to a significant performance drop, showing their efficacy.

\begin{figure}[t]
  \centering
  \includegraphics[width=1\linewidth]{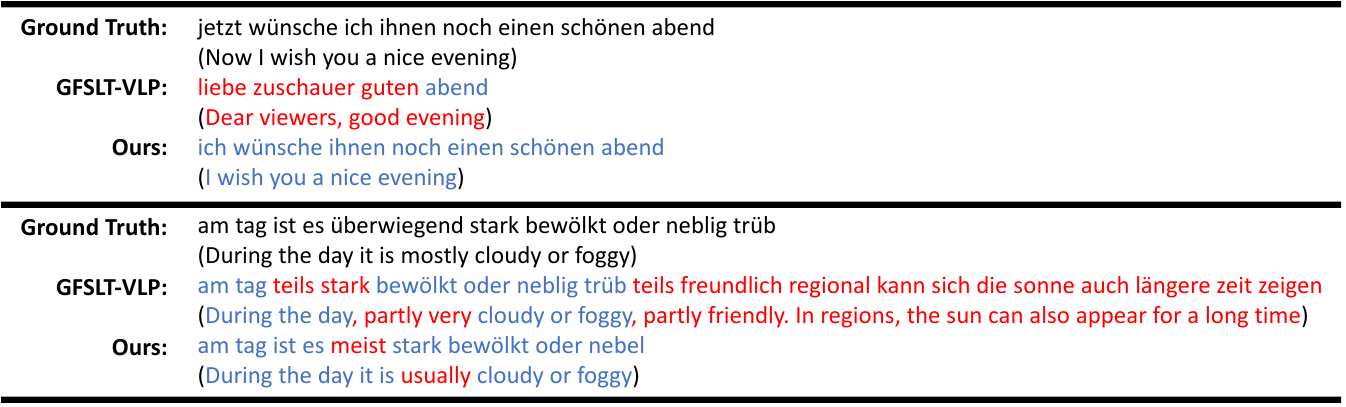}
  \vspace{-8mm}
  \caption{Visualization of translation results. 
  Correct translations are in \textcolor{blue}{blue} while the wrong translations are in \textcolor{red}{red}.   
  }
  \label{fig:ab_qualitative_result}
  \vspace{-5mm}
\end{figure}

\noindent  \textbf{Qualitative Results.}
We present two sample translations generated by our SignLLM and the current state-of-the-art (GFSLT-VLP \cite{SLT:glossfree}) in Fig.~\ref{fig:ab_qualitative_result} for qualitative analysis. 
In the first sample (top), our model produces a highly accurate translation, whereas \cite{SLT:glossfree} inaccurately represents the semantic information. 
In the second sample (bottom), our model successfully preserves sentence semantics, while \cite{SLT:glossfree} introduces a translation error, resulting in redundant and erroneous information.
These examples qualitatively show our SignLLM's efficacy in producing accurate translations.

\section{Conclusion}

We present SignLLM, a novel framework to harness off-the-shelf and frozen LLMs for SLT.
SignLLM imparts language-like characteristics to sign video representations through the VQ-Sign and CRA modules, and a sign-text alignment loss improves semantic compatibility.
We empirically observe that applying our SignLLM leads to state-of-the-art gloss-free results on two popular SLT benchmarks.

\footnotesize{
\noindent
\textbf{Acknowledgements.}
This project is supported by the Ministry of Education, Singapore, under the AcRF Tier 2 Projects (MOE-T2EP20222-0009 and MOE-T2EP20123-0014), National Research Foundation Singapore under its AI Singapore Programme (AISG-100E-2023-121).
}



\end{document}